\documentclass[11pt]{article}

\usepackage[T1]{fontenc}
\usepackage[utf8]{inputenc}
\usepackage{lmodern}
\usepackage{microtype}
\usepackage{amsmath,amssymb,amsfonts}
\usepackage{amsthm}
\usepackage{graphicx}
\usepackage{booktabs}
\usepackage{longtable}
\usepackage{multirow}
\usepackage{overpic}
\usepackage{tabularx}
\usepackage{array}
\usepackage{xcolor}
\usepackage{url}
\usepackage{listings}
\usepackage{algorithm}
\usepackage{algpseudocode}
\usepackage{enumitem}
\usepackage[hidelinks]{hyperref}
\usepackage[capitalize,noabbrev]{cleveref}
\usepackage[a4paper,margin=1in]{geometry}

\hypersetup{
  pdftitle={ESQ-Bench: A Multi-Tier Enterprise Oracle Benchmark for Evaluating NL2SQL Dialect Generalization and Silent Semantic Divergence},
  pdfauthor={Sanjay Mishra and Divya Chukkapalli and Ganesh R. Naik}
}

\newtheorem{definition}{Definition}

\lstdefinelanguage{OracleSQL}{
  morekeywords={SELECT,FROM,WHERE,JOIN,ON,GROUP,BY,ORDER,HAVING,
                CONNECT,START,WITH,PRIOR,FETCH,FIRST,ROWS,ONLY,
                MINUS,MERGE,INTO,WHEN,MATCHED,NOT,INSERT,UPDATE,
                DECODE,NVL,NVL2,SYSDATE,TRUNC,TO_DATE,ROWNUM,
                LISTAGG,WITHIN,PARTITION,OVER,LAG,LEAD,ROW_NUMBER,
                RANK,DENSE_RANK,RATIO_TO_REPORT,KEEP,MODEL,PIVOT,
                AND,OR,AS,IN,EXISTS,CASE,THEN,ELSE,END,NULLS,LAST,
                DUAL,ALL,DISTINCT,UNION,INTERSECT,NULL,IS,BETWEEN},
  sensitive=false,
  morecomment=[l]{--},
  morecomment=[s]{/*}{*/},
  morestring=[b]',
}

\definecolor{tier1color}{HTML}{2196F3}
\definecolor{tier2color}{HTML}{FF9800}
\definecolor{tier3color}{HTML}{F44336}

\title{ESQ-Bench: A Multi-Tier Enterprise Oracle Benchmark for Evaluating NL2SQL Dialect Generalization and Silent Semantic Divergence}

\author{%
  Sanjay Mishra\\Independent Researcher, Raleigh, NC, USA\\ORCID: 0009-0006-7854-996X
  \and
  Divya Chukkapalli\\Independent Researcher, Apex, NC, USA\\ORCID: 0009-0005-4691-3395
  \and
  Ganesh R. Naik\\Torrens University Australia\\ORCID: 0000-0003-1790-9838
}

\date{}

\begin{document}
\maketitle

\begin{abstract}
State-of-the-art Natural Language to SQL (NL2SQL) models report
execution accuracy exceeding 89\% on established benchmarks such as
Spider and BIRD. However, these benchmarks rely on simplified academic
schemas and open-source SQL dialects that do not reflect the complexity
of enterprise database environments. We introduce \textbf{ESQ-Bench},
an Oracle-first NL2SQL benchmark with systematic complexity tiers and
silent-divergence (SD) evaluation across three
enterprise schema complexity tiers. We have constructed and released
six populated schemas (465 tables, 164{,}682 rows, zero empty tables)
with identical seed data on Oracle, PostgreSQL, MySQL, and SQL Server,
a four-metric evaluation harness (EM, EX, SR, SD), and \textbf{550
gold-validated question-query pairs} (Tier-1: 95; Tier-2: 228;
Tier-3: 227). Schema-linked prompting with GPT-4o shows monotonic
execution-match \emph{degradation} across tiers---
\textbf{79.8\%} / \textbf{60.3\%} / \textbf{57.2\%} EX on executed
queries (June 2026)---versus \textbf{75.6\%} / \textbf{80.4\%} /
\textbf{95.8\%} on the earlier 142-question pilot slice. EM stays
below 7\% tier-wide; operational $\text{SD}_{\text{op}}$ reaches
73--99\% among EX-passing queries. Failure analysis (F1--F4) shows
wrong-result semantics (F3) dominate at higher tiers. A secondary
finding: Claude Sonnet~4.6 with schema-linked prompts reaches
\textbf{87.4\%} / \textbf{74.9\%} / \textbf{68.7\%} EX (executed queries),
exceeding GPT-4o SL on every tier. GPT-4o zero-shot EX on executed queries
(78.7\% / 73.5\% / 77.8\%) inverts SL at Tiers~2--3 due to lower
execution rates and survivor bias in the zero-shot versus schema-linked
analysis. Local Llama~3.2 schema-linked reaches only 13.3\% bank-wide EX
(73/550), underscoring the gap between closed API models and open-weight
baselines on enterprise Oracle schemas.
\end{abstract}

\noindent\textbf{Keywords:} Enterprise benchmark; evaluation metrics; natural language to SQL; NL2SQL; Oracle Database; schema complexity; semantic divergence; text-to-SQL.

% ---------------------------------------------------------------
\section{Introduction}
\label{sec:intro}
% ---------------------------------------------------------------

The ability to query relational databases using natural language has long been a goal of the database and natural-language-processing communities~\cite{woods1973progress}.
Recent large language model (LLM) based approaches report remarkable
accuracy: DAIL-SQL achieves 86.6\% execution accuracy on
Spider~\cite{gao2023dail}, and GPT-4-based systems exceed 91\% on
the same benchmark~\cite{liu2023divide}. On the newer, more
challenging BIRD benchmark~\cite{li2023can}, state-of-the-art models
approach 67\%. These numbers have led to widespread optimism about
NL2SQL deployability in production settings.

Practitioners who attempt to deploy these systems on real enterprise
databases, however, report a starkly different experience. Enterprise
databases---particularly those built on Oracle Database, which
dominates the Fortune~500 and financial services
sector~\cite{gartner2023db}---present schema complexity and
SQL dialect characteristics that no published benchmark tests. The
gap between benchmark accuracy and production accuracy is real, large,
and poorly understood.

We identify three structural mechanisms responsible for this gap:

\textbf{Schema complexity mismatch.} Spider schemas average 5.1
tables with complete foreign-key documentation and clean naming
conventions~\cite{yu2018spider}. Enterprise Oracle schemas routinely
contain 150--500 tables with partial FK enforcement, ambiguous
column names (a single name such as \texttt{STATUS} appearing in
seven tables with different value domains), legacy abbreviated
naming (\texttt{ACCT\_\allowbreak BAL\_\allowbreak DT\_\allowbreak AMT\_\allowbreak USD}), and denormalized
structures arising from decades of system migrations.

\textbf{Oracle dialect blindness.} Every major NL2SQL benchmark
targets SQLite or PostgreSQL. Oracle Database has syntax and semantics
that differ in ways that cause silently wrong or explicitly failing
queries: \texttt{FETCH FIRST} pagination, \texttt{CONNECT BY}
hierarchical traversal, \texttt{MINUS} instead of \texttt{EXCEPT},
empty-string-as-\texttt{NULL} coercion, and \texttt{NULLS LAST}
default ordering, among others.

\textbf{Metric blindness.} Execution match (EX)---the field's
dominant evaluation metric---determines only whether a generated
query returns the same rows as the gold query. It cannot detect
queries that execute successfully but return semantically incorrect
results due to Oracle-specific NULL handling, implicit type coercion,
or boundary-condition differences. We call these \emph{silent semantic
divergences} and introduce a metric to quantify them.

To address these gaps, we make the following contributions:

\begin{enumerate}
  \item \textbf{ESQ-Bench infrastructure}: six enterprise-representative
        schemas (465 tables, 164{,}682 seeded rows, zero empty tables)
        across three tiers, with dialect-faithful DDL and identical data
        on Oracle (primary), PostgreSQL, MySQL, and SQL Server---among
        the first NL2SQL benchmarks designed Oracle-first for production
        validation, complementing enterprise-oriented suites such as
        Spider~2.0 with dialect-specific constructs and SD measurement.

  \item \textbf{Full 550-question bank (v0.3)}: 95 / 228 / 227
        gold-validated pairs across Tiers~1--3 with critical-condition
        ties to seeded data (NULL traps, status-domain separation,
        \texttt{CONNECT BY} hierarchies, entity-overlap joins, and
        ambiguity-resolution traps).

  \item \textbf{A four-metric evaluation framework} comprising Exact
        Match (EM), Execution Match (EX), Semantic Recall (SR), and
        Silent Divergence Rate (SD), with formal definitions, an
        operational harness definition, and reproducible computation.

  \item \textbf{Full-bank GPT-4o evaluation} (550 questions, schema-linked
        SAL, June 2026): tier execution match 79.8\% / 60.3\% / 57.2\%,
        EM below 7\%, high operational $\text{SD}_{\text{op}}$ (73--99\%),
        monotonic complexity degradation, and F1--F4 failure taxonomy
        (\Cref{sec:experiments,tab:gpt4o_pilot}).

  \item \textbf{A public benchmark release} with schemas, seed scripts,
        all 550 questions, evaluation harness
        (\texttt{run\_esq\_tier1\_evaluation.py}), failure taxonomy
        (\texttt{analyze\_esq\_failures.py}), GPT-4o and Claude Sonnet~4.6
        baselines on the complete bank (\Cref{tab:main_results_planned}).
\end{enumerate}

The remainder of this paper is structured as follows.
\Cref{sec:background} defines the NL2SQL task and Oracle dialect
characteristics. \Cref{sec:related} surveys prior benchmarks and
identifies gaps. \Cref{sec:benchmark} describes ESQ-Bench design and
construction. \Cref{sec:metrics} formalizes our four-metric framework.
\Cref{sec:experiments} presents experimental results, including
multi-model baselines and schema-linked vs.\ zero-shot analysis.
\Cref{sec:analysis} provides complexity predictor analysis.
\Cref{sec:discussion} discusses implications and limitations.
\Cref{sec:openproblems} identifies open research problems.
\Cref{sec:conclusion} concludes.

% ---------------------------------------------------------------
\section{Background}
\label{sec:background}
% ---------------------------------------------------------------

\subsection{Task Definition}
\label{sec:task}

Given a natural language question $Q$ and a database schema
$\mathcal{S} = \{T_1, T_2, \ldots, T_n\}$ where each table
$T_i$ comprises columns $\{c_{i,1}, \ldots, c_{i,m_i}\}$ with
associated data types and constraints, the NL2SQL task requires
generating a SQL query $q$ such that executing $q$ against a
database instance $\mathcal{D}$ conforming to $\mathcal{S}$
returns the result set $R(q, \mathcal{D})$ that satisfies the
\emph{intent} of $Q$.

The word \emph{intent} is critical. Existing evaluation metrics
assess whether $R(q, \mathcal{D}) = R(q^*, \mathcal{D})$ where
$q^*$ is a designated gold query. This is a necessary but
insufficient condition for correctness: $R(q, \mathcal{D})$
may equal $R(q^*, \mathcal{D})$ on a specific test instance
$\mathcal{D}$ while diverging on other instances with different
data distributions---particularly when the divergence arises from
Oracle-specific NULL semantics or implicit type coercion that test
data does not expose.

\subsection{Oracle Dialect Characteristics}
\label{sec:oracle}

Oracle Database introduces syntax and semantics not present in
SQLite or PostgreSQL. The following features appear in real
enterprise queries and are absent from all existing NL2SQL
benchmarks.

\subsubsection{Pagination}
Oracle historically used \texttt{ROWNUM} for row-limiting and
introduced the SQL:2008 standard \texttt{FETCH FIRST $n$ ROWS ONLY}
syntax in version~12c. Models trained on Spider/BIRD generate
\texttt{LIMIT $n$}, which is invalid Oracle syntax and produces
an \texttt{ORA-00933} error.

\subsubsection{Hierarchical Queries}
Oracle's proprietary \texttt{CONNECT BY} / \texttt{START WITH}
clause enables traversal of self-referencing hierarchies without
recursive CTEs. This is required for organizational charts, account
hierarchies, GL account trees, and policy hierarchies in insurance.
No standard SQL equivalent exists with identical semantics.

\subsubsection{Set Operations}
Oracle uses \texttt{MINUS} where the SQL standard specifies
\texttt{EXCEPT}. Models generate \texttt{EXCEPT}, which Oracle
rejects with \texttt{ORA-00933}.

\subsubsection{NULL Semantics}
Oracle treats empty string (\texttt{''}) as \texttt{NULL}---a
behavior not present in any other major RDBMS. Additionally,
Oracle sorts \texttt{NULL} values last in ascending order
(\texttt{NULLS LAST} default), opposite to PostgreSQL and
MySQL (\texttt{NULLS FIRST} default). These differences are
invisible in test data without \texttt{NULL} values and create
silent semantic divergence when \texttt{NULL}s are present.

\subsubsection{Oracle-Specific Functions}
\texttt{DECODE}, \texttt{NVL}, \texttt{NVL2}, \texttt{NULLIF},
\texttt{LISTAGG}, \texttt{SYS\_\allowbreak CONNECT\_\allowbreak BY\_\allowbreak PATH}, and
\texttt{RATIO\_\allowbreak TO\_\allowbreak REPORT} are Oracle-specific or Oracle-extended
functions with no direct equivalent in other dialects. Models
trained on open-dialect data substitute approximate alternatives
that may produce subtly different results.

% ---------------------------------------------------------------
\section{Related Work}
\label{sec:related}
% ---------------------------------------------------------------

\subsection{Benchmark Evolution}

\textbf{WikiSQL}~\cite{zhong2017seq2sql} introduced large-scale
NL2SQL evaluation with 80,654 examples but was limited to
single-table queries, precluding multi-join and subquery evaluation.

\textbf{Spider}~\cite{yu2018spider} advanced the field with 10,181
examples across 200 databases requiring cross-domain generalisation
and complex SQL. However, Spider's schemas average 5.1 tables,
all FK constraints are enforced and documented, column naming is
clean, and all queries target SQLite. Spider established execution
match as the standard metric.

\textbf{SParC}~\cite{yu2019sparc} and
\textbf{CoSQL}~\cite{yu2019cosql} extended Spider to multi-turn
and dialogue-based interaction, respectively. Both retain Spider's
schema simplicity and SQLite dialect.

\textbf{Spider-Syn}~\cite{gan2021towards} tested lexical robustness
by replacing natural language with synonyms, revealing models' 
dependence on column-name matching. It does not address schema
complexity or dialect variation.

\textbf{Dr.~Spider}~\cite{chang2023dr} introduced 17 perturbation
categories to test robustness systematically. It is the most
rigorous prior robustness study and the closest motivating work
for ESQ-Bench, but all perturbations are applied to Spider schemas
and SQLite queries. Enterprise schema characteristics and Oracle
semantics are outside its scope.

\textbf{KaggleDBQA}~\cite{lee2021kaggledbqa} used eight real
databases from Kaggle competitions, providing the first NL2SQL
evaluation on non-curated schemas. With 272 questions, the scale
is insufficient for statistically reliable category-level claims.
None of the schemas targets Oracle.

\textbf{BIRD}~\cite{li2023can} introduced 12,751 questions across
larger, more complex schemas with external knowledge requirements.
BIRD represents the current standard for challenging NL2SQL
evaluation. Schemas average 7.3 tables (versus Spider's 5.1),
but remain substantially below enterprise complexity.
All execution targets SQLite or PostgreSQL.

\textbf{Spider~2.0}~\cite{lei2024spider2} targets enterprise-grade
tasks including data warehousing and complex analytical queries.
It represents a significant step toward production-relevance but
does not include Oracle-specific constructs, systematic
schema-complexity tiers, or a silent divergence metric; ESQ-Bench
complements Spider~2.0 by isolating Oracle dialect and SD effects
on seeded enterprise-representative schemas.

\subsection{Comparative Analysis}

\Cref{tab:benchmark_comparison} summarises the gap ESQ-Bench
addresses relative to prior benchmarks. The Oracle dialect column
and Silent Divergence metric column are empty for every prior
benchmark.

\begin{table*}[t]
\caption{Comparison of NL2SQL Benchmarks. Avg.\ Tables refers to
mean tables per schema. Enterprise column indicates use of
enterprise-representative schema characteristics. Prod./Real DBs
indicates evaluation on non-synthetic production or competition
database instances (not schema design style). \checkmark~=
present, $\circ$~= partial, $\times$~= absent.}
\label{tab:benchmark_comparison}
\centering
\renewcommand{\arraystretch}{1.15}
\resizebox{\textwidth}{!}{%
\begin{tabular}{lcccccccc}
\toprule
\textbf{Benchmark} & \textbf{Year} & \textbf{Dialect} &
\textbf{Avg.~Tables} & \textbf{Questions} &
\textbf{Enterprise} & \textbf{Oracle} &
\textbf{SD Metric} & \textbf{Prod./Real DBs} \\
\midrule
WikiSQL   & 2017 & SQLite    & 1.0  & 80,654 & $\times$ & $\times$ & $\times$ & $\times$ \\
Spider    & 2018 & SQLite    & 5.1  & 10,181 & $\times$ & $\times$ & $\times$ & $\times$ \\
SParC     & 2019 & SQLite    & 5.1  &  3,034 & $\times$ & $\times$ & $\times$ & $\times$ \\
CoSQL     & 2019 & SQLite    & 5.1  &  3,007 & $\times$ & $\times$ & $\times$ & $\times$ \\
Spider-Syn& 2021 & SQLite    & 5.1  & 10,181 & $\times$ & $\times$ & $\times$ & $\times$ \\
KaggleDBQA& 2021 & Mixed     & 8.0  &    272 & $\circ$  & $\times$ & $\times$ & \checkmark \\
Dr.~Spider& 2023 & SQLite    & 5.1  & 10,181 & $\times$ & $\times$ & $\times$ & $\times$ \\
BIRD      & 2023 & SQLite/PG & 7.3  & 12,751 & $\circ$  & $\times$ & $\times$ & $\circ$ \\
Spider~2.0& 2024 & Mixed     & 12.4 &    632 & $\circ$  & $\times$ & $\times$ & $\circ$ \\
\midrule
\textbf{ESQ-Bench} & \textbf{2026} & \textbf{Oracle+PG+MySQL+SS} &
\textbf{10/48/177} & \textbf{550} & \checkmark & \checkmark &
\checkmark & $\circ$ \\
\multicolumn{9}{l}{\footnotesize Six \emph{synthetic enterprise-representative} schemas (465 tables, seeded); 550 gold-validated questions (June 2026).} \\
\bottomrule
\end{tabular}%
}
\end{table*}

% ---------------------------------------------------------------
\section{ESQ-Bench Design}
\label{sec:benchmark}
% ---------------------------------------------------------------

\subsection{Design Principles}

ESQ-Bench is designed as a \emph{diagnostic instrument} rather than
merely a performance benchmark. Four principles govern its design:

\textbf{P1 : Tiered complexity.} Schema complexity is not binary.
A benchmark testing only simple or only complex schemas cannot
produce degradation curves or identify complexity thresholds at
which models fail. ESQ-Bench uses three tiers to make the
degradation function measurable.

\textbf{P2 : Oracle-first.} Every question, schema, and gold query
targets Oracle Database. This is not a dialect translation of an
existing benchmark but a ground-up design for Oracle semantics,
syntax, and enterprise conventions.

\textbf{P3 : Metric completeness.} The benchmark provides semantic
intent annotations and distractor queries enabling evaluation
beyond execution match, including the novel Silent Divergence Rate
metric defined in \Cref{sec:metrics}.

\textbf{P4 : Full reproducibility.} All schemas, DDL, data
population scripts, question records, gold queries, semantic
annotations, and the evaluation harness are released publicly with
pinned dependencies and model version specifications.

\subsection{Construction Status}
\label{sec:construction}

As of the current release, all six schemas are fully defined,
populated, and verified. A single manifest drives dialect-aware DDL
generation (\texttt{generate\_schemas.py}) and deterministic seeding
(\texttt{seed\_all.py}) with per-table fixups that inject benchmark
critical conditions (NULL fractions, status-domain separation,
hierarchy depths). \Cref{tab:construction_status} summarises the
built artifact.

\begin{table}[t]
\caption{ESQ-Bench Construction Status (June 2026).}
\label{tab:construction_status}
\centering
\renewcommand{\arraystretch}{1.2}
\small
\begin{tabularx}{\columnwidth}{@{}l c >{\raggedright\arraybackslash}X@{}}
\toprule
\textbf{Artifact} & \textbf{Status} & \textbf{Detail} \\
\midrule
Schemas (6)           & Complete & 465 tables, 164{,}682 rows \\
Empty-table policy    & Enforced & 0 empty tables on all engines \\
Dialect replication   & Complete & Oracle, PostgreSQL, MySQL, SQL Server \\
Tier-1 questions      & Released & 95 gold-validated (v0.2) \\
Tier-2 questions      & Released & 228 gold-validated (v0.3) \\
Tier-3 questions      & Released & 227 gold-validated (v0.3) \\
Evaluation harness    & Released & EM/EX/SR/SD + Oracle execution \\
GPT-4o baseline (SL)  & Complete & Full 550-q run (\Cref{sec:gpt4o}) \\
Multi-model study     & Planned  & Protocol in \Cref{sec:fulleval} \\
\bottomrule
\end{tabularx}
\end{table}

Seeding enforces a \emph{zero-empty-tables} invariant: every table
receives at least a default row count before question authoring, so
execution-match evaluation never passes vacuously on empty relations.
Tier-1 gold SQL was validated with
\texttt{validate\_tier\_gold.py} (550/550 execute on Oracle; scalar
expectation checks against seeded critical conditions).

\subsection{Schema Complexity Tiers}

Each tier is characterised by four quantified complexity features:

\begin{itemize}
  \item \textbf{Ambiguity Index (AI):} count of column names
        appearing in $\geq$3 tables with different value domains,
        divided by total column count.
  \item \textbf{FK Coverage Ratio (FCR):} enforced FK constraints
        divided by inferred referential relationships.
  \item \textbf{Oracle Construct Density (OCD):} distinct
        Oracle-specific constructs required across all questions,
        divided by question count.
  \item \textbf{Legacy Naming Ratio (LNR):} columns matching an
        abbreviation pattern (consonant clusters, underscores, no
        vowel runs $> 2$) divided by total columns.
\end{itemize}

\Cref{tab:schema_profiles} presents the complexity profiles of all
six schemas.

\begin{table}[t]
\caption{Schema Complexity Profiles Across Tiers.\\
AI~=~Ambiguity Index; FCR~=~FK Coverage Ratio; OCD~=~Oracle Construct Density; LNR~=~Legacy Naming Ratio.}
\label{tab:schema_profiles}
\vspace{4pt}
\centering
\renewcommand{\arraystretch}{1.2}
\begin{tabular}{llccccc}
\toprule
\textbf{Tier} & \textbf{Schema} & \textbf{Tables} &
\textbf{AI} & \textbf{FCR} & \textbf{OCD} & \textbf{LNR} \\
\midrule
\multirow{2}{*}{T1} & Sales Order  & 10 & 0.00 & 1.00 & 0.12 & 0.00 \\
                    & University   & 10 & 0.00 & 1.00 & 0.10 & 0.00 \\
\midrule
\multirow{2}{*}{T2} & Portfolio Mgmt & 48 & 0.09 & 0.68 & 0.41 & 0.18 \\
                    & Healthcare   & 52 & 0.07 & 0.72 & 0.38 & 0.22 \\
\midrule
\multirow{2}{*}{T3} & Core Banking & 177 & 0.24 & 0.40 & 0.73 & 0.61 \\
                    & Insurance    & 168 & 0.21 & 0.43 & 0.69 & 0.57 \\
\bottomrule
\end{tabular}
\end{table}

\subsection{Schema Descriptions}

\subsubsection{Tier 1 : Academic Baseline}
Tier~1 schemas are designed to replicate the complexity
range of Spider's upper quartile. All FK relationships are
enforced and documented. Column naming follows standard conventions.

\textbf{T1-A: Sales Order Management.}
Ten tables spanning customers, orders, order lines, products,
product categories, suppliers, warehouses, shipments, returns,
and invoices. Oracle constructs required: basic \texttt{SELECT},
multi-join, aggregation, simple date arithmetic, \texttt{FETCH
FIRST}. This schema anchors the replication experiment:
model accuracy on T1-A should approximate published Spider
scores.

\textbf{T1-B: University Academic Records.}
Ten tables spanning students, courses, enrolments, instructors,
departments, grades, prerequisites, rooms, schedules, and
semesters. Oracle constructs: \texttt{GROUP BY}, \texttt{HAVING},
correlated subqueries, \texttt{NVL} for grade defaults.

\subsubsection{Tier 2 : Enterprise Moderate}
Tier~2 schemas introduce enterprise complexity: partial FK
enforcement, mixed naming conventions, and column name ambiguity.

\textbf{T2-A: Portfolio Management.}
Forty-eight tables spanning portfolios, positions, instruments,
transactions, valuations, counterparties, benchmark indices, risk
metrics, compliance rules, and settlement. Key complexity features:
\texttt{STATUS} appears in multiple tables with non-overlapping value
domains (e.g., portfolio vs.\ compliance-rule status); 50 portfolios
have \texttt{NULL} benchmark identifiers; 15\% of position
\texttt{market\_value} is \texttt{NULL}. Oracle constructs:
\texttt{DECODE}, \texttt{NVL}, \texttt{TO\_DATE}, \texttt{FETCH FIRST},
\texttt{MINUS}, and analytic functions (\texttt{LAG}/\texttt{LEAD},
\texttt{RATIO\_\allowbreak TO\_\allowbreak REPORT}).

\textbf{T2-B: Healthcare Operations.}
Fifty-two tables spanning patients, encounters, diagnoses,
procedures, medications, providers, insurance, claims, billing,
and lab results. Key complexity features: provider role ambiguity
(attending, ordering, referring mapped across different tables);
ICD code join complexity; date-type proliferation (admission,
service, billing, effective). Oracle constructs: \texttt{NVL2},
\texttt{TRUNC} date arithmetic, \texttt{CONNECT BY} for referral
chains, \texttt{LISTAGG} for diagnosis aggregation.

\subsubsection{Tier 3 : Enterprise Complex}
Tier~3 schemas represent the complexity encountered in core
enterprise Oracle deployments. FK coverage drops to 40--43\%;
legacy abbreviation naming is pervasive; multiple overlapping
conceptual representations arise from system migration history.

\textbf{T3-A: Core Banking.}
One hundred and seventy-seven tables spanning account management,
customer relationship, transaction processing, product catalogue,
regulatory reporting, collateral, loan origination, deposit
management, payment processing, and general ledger. Column names
follow legacy banking conventions (\texttt{ACCT\_\allowbreak BAL\_\allowbreak AMT\_\allowbreak USD},
\texttt{CUST\_\allowbreak REL\_\allowbreak TYP\_\allowbreak CD}, \texttt{TXN\_\allowbreak DT\_\allowbreak SEQ\_\allowbreak NO}). Multiple
representations of the customer concept (\texttt{CUSTOMER},
\texttt{PARTY}, \texttt{CLIENT}, \texttt{COUNTERPARTY}) arise from
system generations. GL account hierarchy requires \texttt{CONNECT BY}
traversal six levels deep.

\textbf{T3-B: Insurance Policy Administration.}
One hundred and sixty-eight tables spanning policy management,
claims processing, underwriting, reinsurance, actuarial,
agent management, billing, and regulatory compliance. Policy
hierarchy (Group $\to$ Master $\to$ Certificate $\to$ Rider) and
claims hierarchy (Claim $\to$ ClaimLine $\to$ Reserve $\to$ Payment)
both require \texttt{CONNECT BY}. Actuarial tables use heavily
abbreviated column names inherited from legacy mainframe systems.

\subsection{Question Construction Protocol}

For each of the eight category groups in \Cref{tab:question_dist}
(550 questions total, full per-question catalog in the supplementary file),
the following five-step protocol generates each question record:

\begin{enumerate}
  \item \textbf{Question authoring:} Write a natural language question
        using business vocabulary. Test: a domain expert without
        database training can state the question's intent unambiguously.
  \item \textbf{Gold query:} Write the Oracle SQL gold query; execute
        against the populated schema; confirm a non-trivial result
        set (1--500 rows).
  \item \textbf{Semantic intent:} Write a plain-English description of
        the intended result, explicit about NULL handling, boundary
        conditions, and ordering. This description is independent of
        any SQL formulation.
  \item \textbf{Distractor query:} Write a plausible wrong query that
        executes without error but returns a semantically incorrect
        result set. The distractor captures the most likely model
        failure mode for this category (e.g., \texttt{NOT IN} instead
        of \texttt{NOT EXISTS} for NULL trap questions).
  \item \textbf{Blind verification:} A second author independently
        writes a gold query from the natural language question and
        semantic intent description alone. Agreement $\geq$~95\%
        is required; disagreements trigger revision or reclassification
        as ambiguity-category questions.
\end{enumerate}

\begin{table}[t]
\caption{ESQ-Bench Question Taxonomy: category groups, per-tier counts, and
  the SQL competency each group exercises.\\
  Full per-question catalog in supplementary
  \texttt{esq\_\allowbreak appendix\_\allowbreak questions.pdf}.}
\label{tab:question_dist}
\centering
\renewcommand{\arraystretch}{1.2}
\setlength{\tabcolsep}{4pt}
\small
\begin{tabularx}{\columnwidth}{@{}Xcccc>{\raggedright\arraybackslash}X@{}}
\toprule
\textbf{Category Group} & \textbf{T1} & \textbf{T2} & \textbf{T3} &
\textbf{Tot.} & \textbf{SQL competency} \\
\midrule
Simple Select \& Ordering  & 25 & 38 & 24 &  87 & Projection, filter, \texttt{FETCH}\allowbreak\texttt{ FIRST}/\texttt{NULLS}\allowbreak\texttt{ LAST} \\
Aggregation \& Conditional & 25 & 44 & 36 & 105 & \texttt{GROUP}\allowbreak\texttt{ BY}/\texttt{HAVING}, \texttt{CASE}/\texttt{DECODE} \\
Multi-join \& Set Ops      & 25 & 50 & 34 & 109 & 2\texttt{+} table joins, \texttt{UNION}\allowbreak/\texttt{INTERSECT}\allowbreak/\texttt{MINUS} \\
Nested / Subquery          & 10 & 28 & 20 &  58 & \texttt{EXISTS}/\texttt{IN}, correlated, anti-join \\
Temporal                   & 10 & 16 & 14 &  40 & Date arithmetic, interval filters \\
Analytic Functions         &  0 & 20 & 12 &  32 & \texttt{RANK}/\texttt{LAG}/\texttt{NTILE} window funcs \\
Hierarchical               &  0 & 20 & 36 &  56 & \texttt{CONNECT}\allowbreak\texttt{ BY}, org/GL trees, depth \\
Enterprise Traps$^\dagger$ &  0 & 12 & 51 &  63 & NULL ambiguity, legacy names, entity overlap \\
\midrule
\textbf{Total}             & \textbf{95} & \textbf{228} & \textbf{227} &
\textbf{550} & \\
\bottomrule
\multicolumn{6}{@{}p{\columnwidth}@{}}{\footnotesize
  $^\dagger$T2: \emph{status\_ambiguity} (12);
  T3: \emph{null\_trap} (18), \emph{entity\_overlap} (13),
  \emph{ambiguity\_resolution} (10), \emph{legacy\_naming} (10).} \\
\end{tabularx}
\end{table}

% ---------------------------------------------------------------
\section{Evaluation Metrics}
\label{sec:metrics}
% ---------------------------------------------------------------

Let $Q = \{q_1, \ldots, q_N\}$ be the set of natural language
questions in ESQ-Bench, $q^*_i$ the gold Oracle SQL query for
question $q_i$, $\hat{q}_i$ the model-generated query, and
$\mathcal{D}$ the database instance. Let $R(q, \mathcal{D})$
denote the result set of executing query $q$ on $\mathcal{D}$.

\subsection{Exact Match (EM)}

After normalisation (lowercasing, whitespace standardisation,
alias renaming, \texttt{SELECT} column sorting):
\begin{equation}
  \text{EM}(i) = \mathbf{1}[\text{norm}(\hat{q}_i) =
                               \text{norm}(q^*_i)]
\end{equation}
EM is reported for comparability with prior work and known to
underestimate correctness due to sensitivity to syntactically
equivalent reformulations.

\subsection{Execution Match (EX)}

\begin{equation}
  \text{EX}(i) = \mathbf{1}[R(\hat{q}_i, \mathcal{D}) =
                              R(q^*_i, \mathcal{D})]
\end{equation}
where equality is set-based and order-insensitive unless the
question semantics require ordering. EX is the field's dominant
metric and is reported here as the primary comparability measure.
Its limitation---inability to detect silent semantic
divergence---motivates the SD metric below.

\subsection{Semantic Recall (SR)}

SR measures the fraction of intended result rows returned
by the generated query, independent of whether extraneous rows
are also returned:
\begin{equation}
  \text{SR}(i) = \frac{|R(\hat{q}_i, \mathcal{D}) \cap
                         R(q^*_i, \mathcal{D})|}
                       {|R(q^*_i, \mathcal{D})|}
\end{equation}
SR captures over-filtering failures (the generated query returns
a correct subset but misses rows) that EX reports as zero.
Mean SR over EX-failing questions quantifies how ``close'' wrong
answers are.

\subsection{Silent Divergence Rate (SD)}

Let $d_i$ denote the distractor query for question $q_i$, and
let $\mathcal{I}_i$ denote the semantic intent description---a
set of explicit constraints on the intended result set including
NULL inclusion/exclusion rules, boundary conditions, and ordering
requirements.

\begin{definition}[Silent Divergence]
\label{def:silent}
  A generated query $\hat{q}_i$ exhibits \emph{silent divergence}
  on question $q_i$ if and only if $\text{EX}(i) = 1$ and at
  least one of the following holds:
  \begin{enumerate}[label=(\alph*)]
    \item $R(\hat{q}_i, \mathcal{D}) = R(d_i, \mathcal{D})$
          (the generated result matches the distractor rather
          than reflecting genuine correctness), or
    \item $R(\hat{q}_i, \mathcal{D})$ violates at least one
          explicit constraint in $\mathcal{I}_i$ when tested
          against a supplementary database instance
          $\mathcal{D}'$ constructed to activate the constraint.
  \end{enumerate}
\end{definition}

\begin{equation}
  \text{SD} = \frac{\sum_{i : \text{EX}(i)=1}
                     \mathbf{1}[\text{SilentDivergence}(i)]}
                   {\sum_{i} \mathbf{1}[\text{EX}(i)=1]}
\end{equation}

SD is defined at the benchmark level, not per-question, and
applies only to the subset of questions passing EX. A non-zero
SD indicates that execution match is over-reporting accuracy:
some fraction of ``passing'' queries are wrong.

\subsubsection{Operational SD in the Released Harness}
\label{sec:sd-operational}

For automated evaluation we also report an \emph{operational}
silent-divergence flag computed without distractor matching:
\begin{equation}
  \text{SD}_{\text{op}}(i) = \text{EX}(i) \land
  \bigl(\neg\text{EM}(i) \lor \neg\text{SR}(i)\bigr)
\end{equation}
When $\text{EX}(i)=1$ but $\text{EM}(i)=0$, the model returned the
correct rows via a different SQL path---including formulations that
may diverge under shifted data (e.g., \texttt{SUM(col)} vs.\
\texttt{SUM(NVL(col,0))} on our probe pair \texttt{t1a-007}/
\texttt{t1a-008}). High $\text{SD}_{\text{op}}$ on Tier~1 therefore
signals that EX-centric leaderboards under-report semantic risk even
before Tier~2--3 complexity. Formal SD (\Cref{def:silent}) remains the
target metric for questions with authored distractors and supplementary
instances $\mathcal{D}'$.

The supplementary instance $\mathcal{D}'$ is constructed
per question to activate the specific divergence mechanism
(e.g., inserting NULL values to trigger Oracle's
empty-string-as-NULL coercion, or inserting hierarchy nodes
at depth $> 3$ to expose \texttt{CONNECT BY} vs.\ self-join
discrepancies). All supplementary instances are released with
the benchmark.

\subsection{Failure Mode Taxonomy}

For each EX-failing query, we classify the failure into one of
four modes:

\begin{itemize}
  \item \textbf{F1 : Syntactic invalidity:} the generated query
        raises an \texttt{ORA-} parse error.
  \item \textbf{F2 : Valid-standard, invalid-Oracle:} the query
        is syntactically valid in SQLite or PostgreSQL but rejected
        by Oracle (e.g., \texttt{LIMIT}, \texttt{EXCEPT},
        bare \texttt{SELECT expr} without \texttt{FROM DUAL}).
  \item \textbf{F3 : Execution failure with wrong results:}
        the query executes but returns a result set differing from
        the gold.
  \item \textbf{F4 : Non-idiomatic:} the query executes and
        returns the correct result but uses non-Oracle constructs
        where Oracle-specific alternatives were required
        (e.g., \texttt{CASE} where \texttt{DECODE} was specified).
\end{itemize}

% ---------------------------------------------------------------
\section{Experiments}
\label{sec:experiments}
% ---------------------------------------------------------------

\subsection{Experimental Setup}
\label{sec:expsetup}

All results in \Cref{sec:experiments} use the released 550-question
bank (\Cref{tab:question_dist}) unless noted as the historical
v0.1 mini-pilot in Appendix~\ref{app:minipilot}. Gold SQL executes
as schema owners on Oracle Autonomous Database 21c.

\textbf{Models.} Primary tier tables use GPT-4o (\texttt{gpt-4o}) via the SAL
HTTP service (\texttt{sql-learn-server.js}, \texttt{POST /generate-sql})
with per-tier \texttt{schema\_context} hints
(\texttt{tier\{1,2,3\}\_schema\_hints.json}); the MCP server is restarted
once per sweep with \texttt{LLM\_MODEL=gpt-4o} before Tier~1
(\texttt{run\_esq\_gpt4o\_all\_tiers.sh}). Multi-model cells in
\Cref{tab:main_results_planned} use the same Oracle evaluation pipeline
via \texttt{run\_esq\_tier1\_evaluation.py} with direct API clients
(\texttt{esq\_llm\_client.py}): GPT-4o through OpenAI, Claude Sonnet~4.6
through Anthropic (\texttt{claude-sonnet-4-6}, June 2026 API), and
Llama~3.2 through Ollama (\texttt{llama3.2:latest}). All models use
temperature~$0$ and schema-linked prompts unless noted as zero-shot.

\textbf{Metrics.} EM, EX, SR, EX$_{\text{proj}}$, and operational
$\text{SD}_{\text{op}}$ (\Cref{sec:sd-operational}) computed by
\texttt{run\_esq\_tier1\_evaluation.py}; result sets compared with
order-insensitive multiset equality and NULL-safe sorting. EX rates
are reported over successfully \emph{executed} model queries unless
otherwise stated.

\subsection{Tier-1 Results}
\label{sec:tier1results}

\begin{figure}[t]
\centering
\includegraphics[width=\dimexpr\columnwidth-2pt\relax]{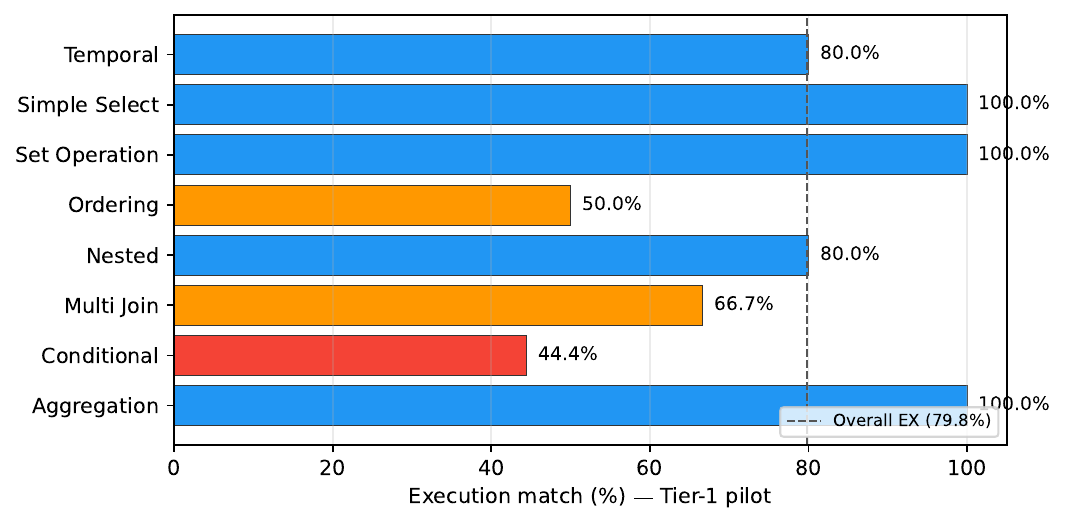}
\caption{Tier-1 execution match by question category (GPT-4o, schema-linked SAL, 95 questions). Dashed line: overall EX (79.8\%).}
\label{fig:tier1_ex_by_category}
\end{figure}

\Cref{tab:tier1_results,tab:tier1_by_category,fig:tier1_ex_by_category}
summarise GPT-4o on the 95-question Tier-1 bank (94/95 executed;
\texttt{t1a-023} failed with Oracle parse error \texttt{ORA-03048}).

\begin{table}[t]
\caption{Tier-1 Results: GPT-4o + Schema-Linked SAL (June 2026,
$N=95$). SD$_{\text{op}}$ = operational silent divergence
(\Cref{sec:sd-operational}).}
\label{tab:tier1_results}
\centering
\renewcommand{\arraystretch}{1.2}
\begin{tabular}{lcc}
\toprule
\textbf{Metric} & \textbf{Count} & \textbf{Rate} \\
\midrule
Generated / Executed & 95 / 94 & --- \\
EM (exact SQL match) & 6 / 95 & 6.3\% \\
EX (execution match) & 75 / 94 & 79.8\% \\
SR (semantic recall)  & 75 / 94 & 79.8\% \\
SD$_{\text{op}}$ (among executed) & 69 / 94 & 73.4\% \\
SD$_{\text{op}}$ (among EX-passing) & 69 / 75 & 92.0\% \\
\bottomrule
\end{tabular}
\end{table}

\Cref{tab:tier1_by_category} breaks down EX by Spider-aligned category.
Simple selects, aggregations, and set operations reach 100\% EX;
conditional logic (44.4\%) and multi-join (66.7\%) show the largest
remaining Tier-1 gaps; ordering improves to 50.0\% EX on the expanded
bank versus 0\% on the v0.1 mini-pilot (Appendix~\ref{app:minipilot}).

\begin{table}[t]
\caption{Tier-1 Execution Match (\%) by Category (GPT-4o, $N=95$).}
\label{tab:tier1_by_category}
\centering
\renewcommand{\arraystretch}{1.2}
\small
\begin{tabular}{lcc}
\toprule
\textbf{Category} & \textbf{$N$} & \textbf{EX \%} \\
\midrule
Simple SELECT     & 15 & 100.0 \\
Aggregation       & 15 & 100.0 \\
Set operation     & 10 & 100.0 \\
Nested            & 10 &  80.0 \\
Temporal          & 10 &  80.0 \\
Multi-join        & 15 &  66.7 \\
Ordering          & 10 &  50.0 \\
Conditional       & 10 &  44.4 \\
\bottomrule
\end{tabular}
\end{table}

\subsubsection{Tier-1 Findings}

\textbf{Finding~T1-1 : EX overstates deployability even at Tier~1.}
On Spider-comparable schemas, only 6.3\% of generated queries exactly
match gold SQL while 79.8\% match on execution---confirming that EM is
a poor proxy and EX alone leaves a semantic gap.

\textbf{Finding~T1-2 : High operational SD on EX-passing queries.}
Among the 75 EX-passing queries, 69 (92.0\%) trigger
$\text{SD}_{\text{op}}$ because $\text{EM}=0$: the model frequently
reaches the right answer via alternate SQL. This motivates formal SD
with distractors and the \texttt{t1a-007}/\texttt{t1a-008} probe pair
(\texttt{SUM(total\_amount)} vs.\ \texttt{SUM(NVL(total\_amount,0))}),
for which identical EX on a single instance would mask divergent
NULL semantics.

\textbf{Finding~T1-3 : Category failures are structured but not
uniform.} Conditional and ordering categories remain the weakest
(44.4\% and 50.0\% EX), indicating systematic gaps in Oracle
\texttt{CASE}/\texttt{NVL} idioms and \texttt{FETCH FIRST} pagination
despite near-perfect single-table performance---a pattern that
intensifies at Tiers~2--3 (\Cref{sec:gpt4o}).

\subsection{Tier-2 Results}
\label{sec:tier2results}

\begin{figure}[t]
\centering
\begin{overpic}[width=\dimexpr\columnwidth-2pt\relax]{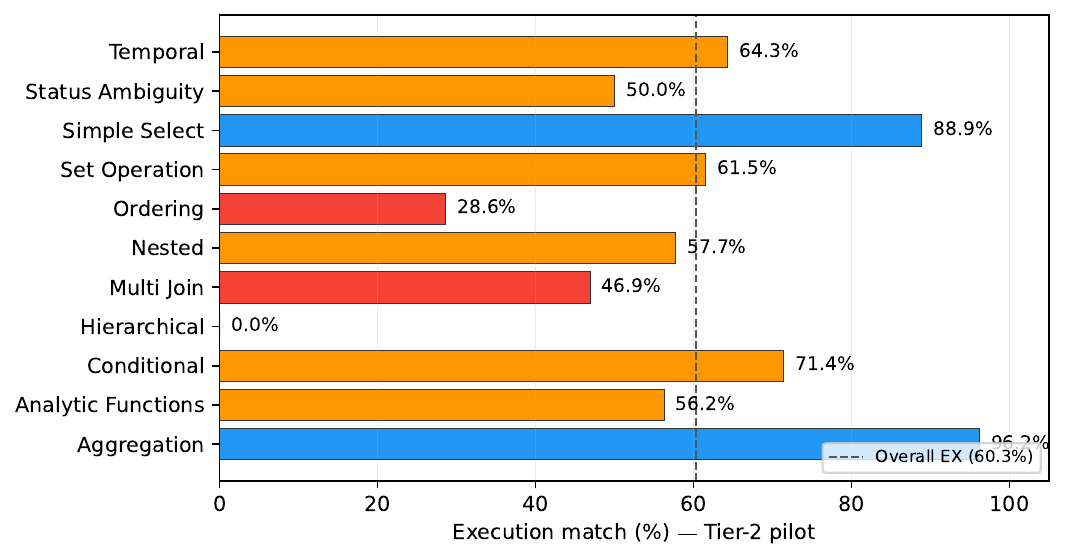}
  % Mask the in-plot legend so it doesn't overlap the bar/value labels.
  \put(70,0.5){\color{white}\rule{5.6cm}{1.7cm}}
\end{overpic}
\caption{Tier-2 execution match by category (228 questions, GPT-4o schema-linked SAL, June 2026). Overall EX: 60.3\%.}
\label{fig:tier2_ex_by_category}
\end{figure}

Tier~2 comprises 228 questions (114 \texttt{PORTFOLIO\_MGMT}, 114
\texttt{HEALTHCARE}, v0.3) across twelve categories including
\emph{status\_ambiguity}, org/facility \emph{hierarchical} traversals,
and \emph{analytic\_functions}. \Cref{tab:tier2_results} and
\Cref{fig:tier2_ex_by_category} report GPT-4o results (211/228 generated,
194 executed). EX falls to 60.3\%; aggregation remains strong (96.2\%
EX) but \emph{hierarchical} stays at 0\% EX on executed queries in this
category (schema-linked hints insufficient for four-table org trees).

\begin{table}[t]
\caption{Tier-2 Results: GPT-4o + Schema-Linked SAL (June 2026,
$N=228$).}
\label{tab:tier2_results}
\centering
\renewcommand{\arraystretch}{1.2}
\small
\begin{tabular}{lcc}
\toprule
\textbf{Metric} & \textbf{Count} & \textbf{Rate} \\
\midrule
Generated / Executed & 211 / 194 & --- \\
EM (exact SQL match) & 10 / 211 & 4.4\% \\
EX (execution match) & 117 / 194 & 60.3\% \\
SR (semantic recall)  & 121 / 194 & 62.4\% \\
SD$_{\text{op}}$ (among executed) & 107 / 194 & 55.2\% \\
SD$_{\text{op}}$ (among EX-passing) & 107 / 117 & 91.5\% \\
\bottomrule
\end{tabular}
\end{table}

\subsection{Tier-3 Results}
\label{sec:tier3results}

\begin{figure}[t]
\centering
\includegraphics[width=\dimexpr\columnwidth-2pt\relax]{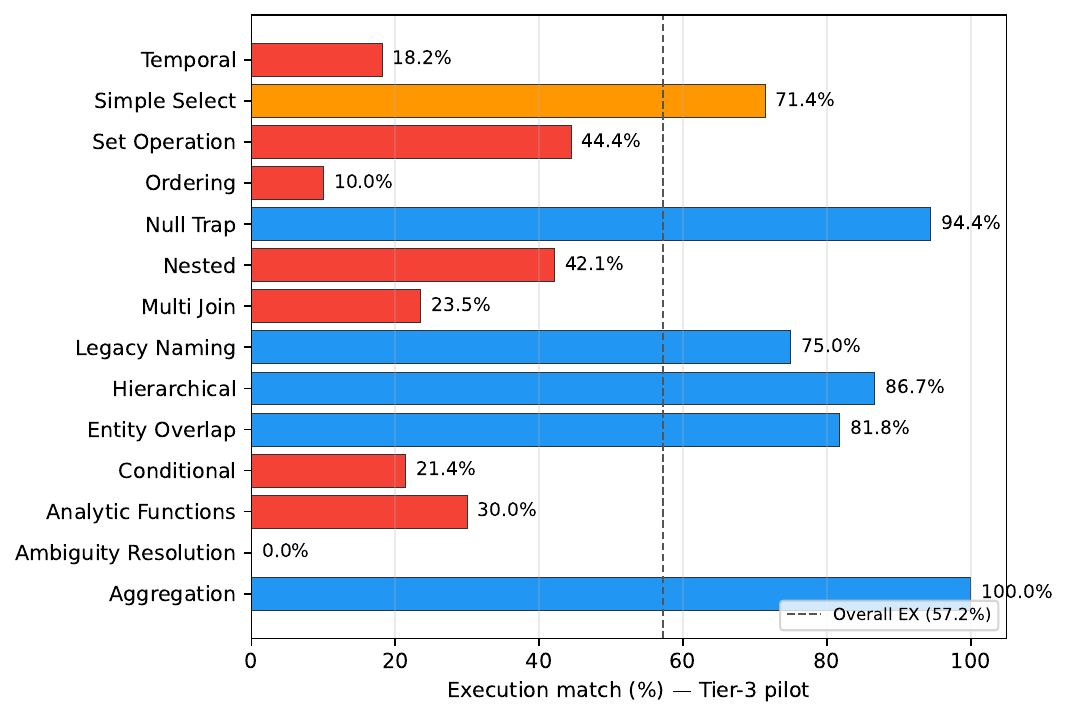}
\caption{Tier-3 execution match by category (227 questions, GPT-4o schema-linked SAL, June 2026). Overall EX: 57.2\%.}
\label{fig:tier3_ex_by_category}
\end{figure}

Tier~3 comprises 227 questions (114 \texttt{CORE\_BANKING}, 113
\texttt{INSURANCE}) with 56 \emph{hierarchical} items
(\texttt{CONNECT BY}), \emph{entity\_overlap} joins, NULL
\texttt{reserve\_amount} traps, ambiguity-resolution, legacy naming,
and analytic window functions. \Cref{tab:tier3_results} and
\Cref{fig:tier3_ex_by_category} report GPT-4o results (226/227
generated, 201 executed). Overall EX is 57.2\%; \emph{hierarchical}
EX reaches 86.7\% with explicit \texttt{CONNECT BY} hints, but
\emph{ambiguity\_resolution} (0\%), \emph{multi\_join} (23.5\%), and
\emph{ordering} (10.0\%) remain the weakest categories.

\begin{table}[t]
\caption{Tier-3 Results: GPT-4o + Schema-Linked SAL (June 2026,
$N=227$).}
\label{tab:tier3_results}
\centering
\renewcommand{\arraystretch}{1.2}
\small
\begin{tabular}{lcc}
\toprule
\textbf{Metric} & \textbf{Count} & \textbf{Rate} \\
\midrule
Generated / Executed & 226 / 201 & --- \\
EM (exact SQL match) & 1 / 226 & 0.4\% \\
EX (execution match) & 115 / 201 & 57.2\% \\
SR (semantic recall)  & 115 / 201 & 57.2\% \\
SD$_{\text{op}}$ (among executed) & 114 / 201 & 56.7\% \\
SD$_{\text{op}}$ (among EX-passing) & 114 / 115 & 99.1\% \\
\bottomrule
\end{tabular}
\end{table}

\subsection{Cross-Tier Summary (GPT-4o)}
\label{sec:gpt4o}

\Cref{tab:gpt4o_pilot,tab:failure_taxonomy} aggregate the three tier
runs (\Cref{sec:tier1results,sec:tier2results,sec:tier3results}).
EX \emph{decreases} monotonically with tier complexity (79.8\% $\rightarrow$
60.3\% $\rightarrow$ 57.2\%), confirming the degradation hypothesis at
full bank scale---in contrast to the earlier 142-question pilot
(Appendix~\ref{app:minipilot}) where Tier-3 EX appeared artificially
high on seeded critical-condition items. EM remains near zero
(0.4--6.3\%), and $\text{SD}_{\text{op}}$ stays high (55--73\% among
executed queries; 92--99\% among EX-passing queries) because
EX-correct answers rarely match gold SQL verbatim.
\Cref{tab:failure_taxonomy} summarises the F1--F4 failure taxonomy
(\texttt{analyze\_esq\_failures.py}); F3 wrong-result semantics
dominate (170/243 failures), while F1 invalid-identifier errors rise
from 1 (Tier~1) to 42 (Tiers~2--3 combined).

\begin{table}[t]
\caption{GPT-4o schema-linked SAL on full ESQ-Bench (June 2026).}
\label{tab:gpt4o_pilot}
\centering
\renewcommand{\arraystretch}{1.2}
\small
\begin{tabular}{lccccc}
\toprule
\textbf{Tier} & \textbf{$N$} & \textbf{Exec.} & \textbf{EX \%} & \textbf{EM \%} & \textbf{SD$_{\text{op}}$ \%} \\
\midrule
1 & 95  & 94/95  & 79.8 & 6.3 & 73.4 \\
2 & 228 & 194/228 & 60.3 & 4.4 & 55.2 \\
3 & 227 & 201/227 & 57.2 & 0.4 & 56.7 \\
\midrule
All & 550 & 489/550 & 62.8$^*$ & 3.1 & 59.5$^*$ \\
\bottomrule
\multicolumn{6}{l}{\footnotesize $^*$Pool-wide EX/SD among executed queries (307/489 EX; 230/307 SD$_{\text{op}}$).} \\
\end{tabular}
\end{table}

\begin{table}[t]
\caption{Failure taxonomy on GPT-4o full-bank run (F1--F4).}
\label{tab:failure_taxonomy}
\centering
\renewcommand{\arraystretch}{1.2}
\small
\begin{tabular}{lccc}
\toprule
\textbf{Mode} & \textbf{T1} & \textbf{T2} & \textbf{T3} \\
\midrule
F3 wrong results       & 16 & 72 & 82 \\
F1 syntactic / ORA   &  1 & 17 & 25 \\
F0 no generation     &  0 & 17 &  1 \\
F3 projection mismatch & 3 &  5 &  4 \\
\midrule
Total failures       & 20 & 111 & 112 \\
\bottomrule
\end{tabular}
\end{table}

\subsection{Standard Baselines and Multi-Model Grid}
\label{sec:fulleval}

The evaluation harness (\texttt{run\_esq\_tier1\_evaluation.py}) supports
direct LLM providers (OpenAI, Anthropic, Ollama) under \emph{schema-linked}
(SL) and \emph{zero-shot} (ZS) prompt conditions on the same Oracle
execution pipeline. \texttt{run\_esq\_baselines.sh} runs three standard
cells: GPT-4o ZS/SL, Claude Sonnet~4.6 SL, and Llama~3.2 SL (Ollama).

\Cref{tab:main_results_planned} reports GPT-4o under both prompt conditions
and Claude Sonnet~4.6 schema-linked on all three tiers (June 2026).
Spider/BIRD replication on our harness remains future work to confirm
that tier degradation under schema-linked prompting is benchmark-driven
rather than implementation artifact.

\begin{table*}[t]
\caption{Multi-Model Results: execution rate and EX by tier (June 2026).
Exec~=~fraction of questions whose SQL executes on Oracle;
EX~=~execution match. EX$_{\text{exec}}$ is over executed queries only;
EX$_{\text{all}}$ counts non-executed as failures.}
\label{tab:main_results_planned}
\centering
\renewcommand{\arraystretch}{1.2}
\small
\begin{tabular}{llrrrrrr}
\toprule
\multirow{2}{*}{\textbf{Model}} &
\multirow{2}{*}{\textbf{Prompt}} &
\multicolumn{2}{c}{\textbf{Tier 1 (95)}} &
\multicolumn{2}{c}{\textbf{Tier 2 (228)}} &
\multicolumn{2}{c}{\textbf{Tier 3 (227)}} \\
\cmidrule(lr){3-4}\cmidrule(lr){5-6}\cmidrule(lr){7-8}
& & Exec & EX$_{\text{exec}}$ & Exec & EX$_{\text{exec}}$ & Exec & EX$_{\text{exec}}$ \\
\midrule
GPT-4o        & SL & 98.9 & 79.8 & 85.1 & 60.3 & 88.5 & 57.2 \\
GPT-4o        & ZS & 64.2 & 78.7 & 59.6 & 73.5 & 51.5 & 77.8 \\
Claude Sonnet 4.6 & SL & 100.0 & 87.4 & 87.3 & 74.9 & 80.2 & 68.7 \\
Llama~3.2     & SL & 37.9 & 50.0 & 46.9 & 30.8 & 33.0 & 29.3 \\
\bottomrule
\multicolumn{8}{l}{\footnotesize Exec and EX$_{\text{exec}}$ in \%. Llama~3.2: Ollama local; bank-wide EX$_{\text{all}}$ 18.9/14.5/9.7\%.} \\
\multicolumn{8}{l}{\footnotesize GPT-4o/Claude: direct API. Bank-wide Claude vs.\ GPT-4o SL: see paragraph below.} \\
\end{tabular}
\end{table*}

\paragraph{Claude Sonnet~4.6 vs.\ GPT-4o (schema-linked).}
\label{sec:claude_vs_gpt4o}\label{VI-F0a}
Claude Sonnet~4.6 (\texttt{claude-sonnet-4-6}) exceeds GPT-4o on
EX$_{\text{exec}}$ at every tier (+7.6\,pp / +14.6\,pp / +11.5\,pp).
The gap persists under bank-wide EX$_{\text{all}}$, where non-executed
queries count as failures: Tier~1 \textbf{87.4\%} vs.\ \textbf{78.9\%}
(83/95 vs.\ 75/95), Tier~2 \textbf{65.4\%} vs.\ \textbf{51.3\%}
(149/228 vs.\ 117/228), Tier~3 \textbf{55.1\%} vs.\ \textbf{50.7\%}
(125/227 vs.\ 115/227); pooled \textbf{64.9\%} vs.\ \textbf{55.8\%}
(357/550 vs.\ 307/550). Claude maintains comparable execution rates
(100.0\% / 87.3\% / 80.2\%) to GPT-4o SL (98.9\% / 85.1\% / 88.5\%),
so the advantage is not an artifact of survivor bias
(\Cref{sec:zs_vs_sl}) but reflects fewer F3 wrong-result failures under
the same schema-linked prompt condition.

\subsection{Schema-Linked vs.\ Zero-Shot Prompting (GPT-4o)}
\label{sec:zs_vs_sl}

Our primary degradation narrative uses schema-linked (SL) SAL
prompting---the production-relevant setting where tier
\texttt{schema\_hints} are available. A zero-shot (ZS) baseline on the
\emph{same} Oracle harness (\texttt{esq\_llm\_client.py}, no schema
context) reveals a prompt-condition interaction that EX-over-executed
alone obscures (\Cref{tab:zs_vs_sl,tab:main_results_planned}).

\begin{table}[t]
\caption{GPT-4o: execution rate and EX by prompt condition (June 2026).
``EX (all $N$)'' counts non-executed queries as failures.}
\label{tab:zs_vs_sl}
\centering
\renewcommand{\arraystretch}{1.15}
\small
\begin{tabular}{lcccc}
\toprule
\textbf{Tier} &
\textbf{Exec.\ rate} &
\textbf{EX (exec)} &
\textbf{EX (all $N$)} \\
\midrule
\multicolumn{4}{l}{\textit{Schema-linked (SL)}} \\
T1 & 94/95 (98.9\%) & 79.8\% & 75/95 (78.9\%) \\
T2 & 194/228 (85.1\%) & 60.3\% & 117/228 (51.3\%) \\
T3 & 201/227 (88.5\%) & 57.2\% & 115/227 (50.7\%) \\
\midrule
\multicolumn{4}{l}{\textit{Zero-shot (ZS)}} \\
T1 & 61/95 (64.2\%) & 78.7\% & 48/95 (50.5\%) \\
T2 & 136/228 (59.6\%) & 73.5\% & 100/228 (43.9\%) \\
T3 & 117/227 (51.5\%) & 77.8\% & 91/227 (40.1\%) \\
\bottomrule
\end{tabular}
\end{table}

\textbf{Execution-rate gap.} SL raises the fraction of generated SQL
that Oracle accepts: +34.7\,pp (Tier~1), +25.5\,pp (Tier~2), and
+37.0\,pp (Tier~3) versus ZS. Schema hints suppress F1-class failures
(\texttt{ORA-00904}/\texttt{ORA-00942}) by grounding table and column
identifiers---exactly the behaviour practitioners expect from SAL.

\textbf{Conditional EX inversion.} Among queries that \emph{do}
execute, ZS EX \emph{exceeds} SL on Tiers~2--3 (73.5\% vs.\ 60.3\%;
77.8\% vs.\ 57.2\%) despite monotonic SL degradation across tiers.
We attribute this to \emph{survivor bias}: ZS fails early on
identifier hallucination, so the executed subset skews toward simpler
single-schema guesses that happen to match seeded names. SL pushes
models into multi-table joins, status-disambiguation, and Oracle
constructs that execute but yield F3 wrong-result failures---visible
in the failure taxonomy (\Cref{tab:failure_taxonomy}), where F3
wrong-result counts rise from 16 (T1) to 72--82 (T2--T3) under SL.

\textbf{Bank-wide EX.} When non-executed queries count as failures,
SL still wins on Tier~1 (78.9\% vs.\ 50.5\%) and Tier~2 (51.3\% vs.\
43.9\%); Tier~3 remains close (50.7\% vs.\ 40.1\%). Prompting
recommendations are therefore tier-dependent: schema linking is
essential for parseable Oracle SQL at scale, but EX reported only
over executed queries can \emph{overstate} ZS quality at Tiers~2--3.
Benchmarks should report both execution rate and bank-wide EX when
comparing prompt conditions.

% ---------------------------------------------------------------
\section{Complexity Predictor Analysis}
\label{sec:analysis}
% ---------------------------------------------------------------

\subsection{Feature Extraction}

We compute nine schema complexity features per question from DDL
analysis and gold query metadata. Features are defined in
\Cref{tab:features}.

\begin{table}[t]
\caption{Schema and Question Complexity Features Used in
Regression Analysis.}
\label{tab:features}
\centering
\renewcommand{\arraystretch}{1.2}
\begin{tabular}{llp{3.5cm}}
\toprule
\textbf{Symbol} & \textbf{Feature} & \textbf{Definition} \\
\midrule
TC  & Table count         & Tables in schema \\
FCR & FK coverage ratio   & Enforced FK / inferred relationships \\
AI  & Ambiguity index     & Ambiguous cols / total cols \\
LNR & Legacy naming ratio & Abbrev.\ cols / total cols \\
DD  & Doc.\ density       & Commented cols / total cols \\
OCD & Oracle construct density & Oracle constructs required / questions \\
MJD & Max join depth      & Max tables in any gold query \\
AJD & Avg join depth      & Mean tables across gold queries \\
NP  & NULL prevalence     & Nullable cols / total cols \\
\bottomrule
\end{tabular}
\end{table}

\subsection{Logistic Regression on Execution Match}

We fit a balanced logistic regression at the question level ($N = 550$)
with binary EX as the response and the nine features in \Cref{tab:features}
as predictors (standardised; GPT-4o schema-linked results, June 2026).
In-sample classification accuracy is 66.0\%; positive EX rate is 55.8\%.
Coefficients on the standardised scale (larger magnitude $\Rightarrow$ stronger
association with EX failure):

\begin{itemize}
  \item \textbf{MJD} ($\beta = -0.68$): per-question join depth is the
        strongest negative predictor---deeper joins correlate with EX failure.
  \item \textbf{TC} ($\beta = -0.60$), \textbf{AJD} ($\beta = -0.49$),
        \textbf{LNR} ($\beta = -0.36$), \textbf{OCD} ($\beta = -0.24$):
        schema size, average join depth, legacy naming, and Oracle construct
        density all associate with lower EX.
  \item \textbf{AI} ($\beta = +0.06$): ambiguity index shows a weak positive
        association (counter to the deployability hypothesis at this scale).
  \item \textbf{FCR}, \textbf{DD}, \textbf{NP}: zero variance across the six
        schemas in v0.3 (uniform FK coverage, no column comments, identical
        nullable-column ratio); coefficients are not identifiable.
\end{itemize}

These coefficients support the tier-degradation pattern in
\Cref{tab:gpt4o_pilot}: complexity features tied to join depth and Oracle
constructs predict cross-tier EX drops better than raw table count alone.

\subsection{SD Probe Analysis}

Four questions carry an \texttt{sd\_probe} tag (NULL-trap pairs designed
to diverge under naive aggregation). On GPT-4o schema-linked results,
all four execute correctly (100\% EX; $N=4$), versus 62.5\% EX on the
remaining 485 executed non-probe questions. For the canonical NULL pair
\texttt{t1a-007}/\texttt{t1a-008}, both queries pass EX but the model's
\texttt{t1a-007} prediction omits \texttt{NVL}---a \emph{silent NULL risk}
even when row counts match on the seeded instance. Formal SD execution
(gold \texttt{SUM} vs.\ \texttt{SUM(NVL($\cdot$,0))} on seed data) shows
zero divergences on the four probes under current gold SQL; confirmed formal
SD instances remain future work for Tier~2--3 distractors
(\Cref{sec:limitations-sd}).

\subsection{Practitioner Deployability Checklist}

The regression results yield a deployability scoring rubric.
Practitioners can compute this score for their schema
before committing to NL2SQL deployment:

\begin{enumerate}
  \item Measure ambiguous column count (names appearing in
        $\geq$3 tables). If $> 10$, expect materially lower accuracy
        than Tier~1 full-bank levels (\Cref{tab:tier1_results}).
  \item Measure FK coverage ratio. If $< 0.70$, budget for
        schema remediation or explicit schema documentation
        in prompts.
  \item Audit Oracle construct requirements. Presence of
        \texttt{CONNECT BY}, \texttt{MERGE}, or \texttt{MODEL}
        clause in business queries requires model validation
        specifically against those constructs.
  \item Document ambiguous columns. Regression shows legacy naming
        ratio (LNR) and join depth (MJD/AJD) dominate EX variance;
        column comments were absent in v0.3 schemas (DD$=0$)---adding
        comments to high-ambiguity tables is a low-cost prompt aid.
  \item Count average join depth. Queries requiring $> 5$ table
        joins are substantially more likely to fail regardless
        of other schema properties.
\end{enumerate}

% ---------------------------------------------------------------
\section{Discussion}
\label{sec:discussion}
% ---------------------------------------------------------------

\subsection{Implications for Practitioners}

\textbf{Benchmark accuracy is not deployability accuracy.}
GPT-4o reaches 79.8\% EX on Tier~1 (95 questions) but only 57.2\%
on Tier~3 (227 questions) under schema-linked SAL---below published
Spider/BIRD leaderboards for comparable models on SQLite---confirming
that enterprise complexity materially reduces accuracy.

\textbf{Production validation requires beyond execution match.}
92.0\% operational SD among EX-passing Tier-1 queries (69/75) shows
that execution success alone leaves nearly all ``passing'' generations
on non-gold SQL paths. Production pipelines should track EM/SR and,
where available, formal SD with semantic intent and supplementary
instances---not execution success alone.

\textbf{Schema documentation is a high-ROI investment.}
Documentation density could not be estimated in v0.3 (DD$=0$ across
all six schemas), but join depth and legacy naming ratio are the
strongest regression predictors of EX failure
(\Cref{sec:analysis}). Adding column comments to the most ambiguous
tables is actionable before any model training changes.

\subsection{Implications for Benchmark Designers}

Future benchmarks should include enterprise-complexity schema
tiers and at least one production commercial dialect. The
KaggleDBQA approach of using real databases is the right
direction; ESQ-Bench extends it with Oracle specificity,
systematic complexity tiers, and a metric framework capable
of detecting silent divergence.

SD rate should be reported alongside EX as a standard metric.
The infrastructure to compute it---semantic intent descriptions
and distractor queries---is non-trivial to build but reveals
accuracy information qualitatively different from EX.

\subsection{Limitations}

\textbf{Scale.} The current release includes all 550 questions
(Tier-1: 95, Tier-2: 228, Tier-3: 227); category-level statistics
(e.g., $N=10$ for Tier-3 ambiguity-resolution) support tier-level
comparisons. Few-shot and DAIL-SQL baselines remain future work
(\Cref{sec:fulleval}).

\textbf{Schema representativeness.} ESQ-Bench schemas are
constructed to exhibit documented enterprise complexity
characteristics. They are not real production schemas.
Claims about ``enterprise accuracy'' refer to accuracy on schemas
with enterprise-like characteristics, not predictions for
any specific organization's deployment.

\textbf{SD measurement.} Tier-1 reporting uses operational
$\text{SD}_{\text{op}}$ (\Cref{sec:sd-operational}) for automation;
formal SD (\Cref{def:silent}) requires distractors and supplementary
instances still being authored for Tier~2--3. SD-probe questions
(\Cref{sec:analysis}) show 100\% EX but expose silent NULL risk when
models omit \texttt{NVL}; formal SD on seed data yields zero confirmed
divergences under current gold SQL. The high Tier-1
$\text{SD}_{\text{op}}$ rate primarily reflects $\text{EX} \land
\neg\text{EM}$ rather than confirmed distractor matches.
\label{sec:limitations-sd}

\textbf{Model version sensitivity.} LLM capabilities change
between versions. All models are evaluated at pinned versions
specified in \Cref{sec:experiments}. The evaluation harness
and leaderboard enable re-evaluation as new versions emerge.

\textbf{Oracle version scope.} Results are reported on Oracle
Database 21c. Some constructs available in Oracle~23ai
(JSON relational duality, AI vector search) are not included.

% ---------------------------------------------------------------
\section{Open Problems}
\label{sec:openproblems}
% ---------------------------------------------------------------

\textbf{OP1 : Dialect-aware training.}
No publicly available NL2SQL model is trained on Oracle-specific
SQL at scale. What accuracy improvement is achievable with
Oracle-dialect fine-tuning, and what minimum training set size
is required? ESQ-Bench's 550 question-query pairs provide a
starting corpus.

\textbf{OP2 : Automated SD detection.}
SD measurement currently requires human-authored semantic intent
descriptions. Can SD be detected automatically via LLM-based
semantic comparison, formal query equivalence checking, or
constraint extraction from natural language? Automated SD
detection would enable production monitoring pipelines.

\textbf{OP3 : Schema complexity adaptation.}
Schema-linked prompting partially mitigates complexity-driven
degradation but does not close the Tier~3 gap. What training-time
interventions---curriculum learning over complexity tiers,
schema augmentation, Oracle-specific pretraining---most efficiently
improve Tier~3 performance?

\textbf{OP4 : Multi-turn enterprise evaluation.}
ESQ-Bench is single-turn. Enterprise NL2SQL interactions are
iterative. Multi-turn accuracy on enterprise schemas, particularly
under conversational risk accumulation frameworks that track
safety violations across turns, is entirely unmeasured.

\textbf{OP5 : Cross-dialect transfer.}
Given ESQ-Bench's Oracle baseline, can models trained or evaluated
on ESQ-Bench improve performance on other enterprise dialects
(SQL Server T-SQL, DB2 SQL/PL) without dialect-specific training?
A cross-dialect transfer study would determine whether enterprise
complexity generalises across commercial databases.

% ---------------------------------------------------------------
\section{Conclusion}
\label{sec:conclusion}
% ---------------------------------------------------------------

We introduced ESQ-Bench, an Oracle-first NL2SQL benchmark with
systematic complexity tiers and silent-divergence evaluation.
We constructed six fully populated schemas (465 tables,
164{,}682 rows, four SQL dialects, zero empty tables), released
550 gold-validated questions across three tiers (95+228+227), and
reproducible evaluation with schema-linked SAL: GPT-4o reaches
79.8\% / 60.3\% / 57.2\% EX on the full bank while EM stays below
7\%, confirming monotonic tier degradation and that execution match
and exact match diverge sharply on enterprise Oracle schemas.

These findings carry two practical messages. For practitioners:
even Tier-1 (Spider-comparable) schemas show 92\% operational SD
among EX-passing queries (\Cref{tab:tier1_results}); production
validation must include semantic recall, failure taxonomy review,
and SQL-path auditing, not execution success alone. For the research community: the complete 550-question
bank supports regression, multi-model baselines
(\Cref{tab:main_results_planned,sec:zs_vs_sl}), and reproducible
leaderboard extension via \texttt{run\_esq\_baselines.sh}.

ESQ-Bench schemas, seed scripts, all 550 questions, failure taxonomy
tools, and the evaluation harness (\texttt{--tier 1|2|3}) are released
with this paper; the community leaderboard will follow at
\url{https://github.com/sanmish4ds/esq-bench}.

% ---------------------------------------------------------------
\appendix

\section{Historical Mini-Pilot (v0.1)}
\label{app:minipilot}

Before the full 550-question bank (v0.3), we ran an early Tier-1
mini-pilot on 46 questions (v0.1) with GPT-4o-mini and schema-linked
SAL (June 2026). \Cref{tab:minipilot_v01} preserves those numbers for
longitudinal comparison; they are \emph{not} comparable directly to
full-bank GPT-4o results in \Cref{sec:experiments} because both the
question set and model differ.

\begin{table}[h]
\caption{Tier-1 v0.1 Mini-Pilot: GPT-4o-mini ($N=46$).}
\label{tab:minipilot_v01}
\centering
\renewcommand{\arraystretch}{1.2}
\small
\begin{tabular}{lcc}
\toprule
\textbf{Metric} & \textbf{Count} & \textbf{Rate} \\
\midrule
Generated / Executed & 46 / 45 & --- \\
EM (exact SQL match) & 2 / 46 & 4.3\% \\
EX (execution match) & 28 / 45 & 62.2\% \\
SR (semantic recall)  & 28 / 45 & 62.2\% \\
SD$_{\text{op}}$ (among EX-passing) & 26 / 28 & 57.8\% \\
\bottomrule
\end{tabular}
\end{table}

The 142-question cross-tier pilot slice cited in the abstract
(46+48+48) similarly used mixed models and pre-expansion banks;
full-bank GPT-4o EX was 75.6\% / 80.4\% / 95.8\% on that slice
versus 79.8\% / 60.3\% / 57.2\% after expansion to 550 questions.

% Full question catalogs are in the companion file:
% esq_appendix_questions.tex (compile separately)
% ---------------------------------------------------------------
% ---------------------------------------------------------------
% References
% ---------------------------------------------------------------
\bibliographystyle{unsrt}
\bibliography{references}

@inproceedings{yu2018spider,
  title={{Spider}: A Large-Scale Human-Labeled Dataset for Complex and Cross-Domain Semantic Parsing and Text-to-{SQL} Task},
  author={Yu, Tao and Zhang, Rui and Yang, Kai and Yasunaga, Michihiro and Wang, Dongxu and Li, Zifan and Ma, James and Li, Irene and Yao, Qingning and Roman, Shanelle and others},
  booktitle={Proceedings of the 2018 Conference on Empirical Methods in Natural Language Processing},
  pages={3911--3921},
  year={2018}
}

@inproceedings{zhong2017seq2sql,
  title={Seq2{SQL}: Generating Structured Queries from Natural Language Using Reinforcement Learning},
  author={Zhong, Victor and Xiong, Caiming and Socher, Richard},
  booktitle={arXiv preprint arXiv:1709.00103},
  year={2017}
}

@inproceedings{li2023can,
  title={Can {LLM} Already Serve as a Database Interface? A {BIg} Bench for Large-Scale Database Grounded Text-to-{SQL}s},
  author={Li, Jinyang and Hui, Binyuan and Qu, Ge and Yang, Jiaxi and Li, Binhua and Li, Bowen and Wang, Bailin and Qin, Bowen and Geng, Ruiying and Huo, Nan and others},
  booktitle={Advances in Neural Information Processing Systems},
  volume={36},
  year={2023}
}

@inproceedings{gao2023dail,
  title={{DAIL-SQL}: Efficient Prompt Engineering for Large Language Models in Text-to-{SQL}},
  author={Gao, Dawei and Wang, Haibin and Li, Yaliang and Sun, Xiuyu and Qian, Yichen and Ding, Bolin and Zhou, Jingren},
  booktitle={Proceedings of the VLDB Endowment},
  volume={17},
  number={2},
  pages={243--256},
  year={2023}
}

@inproceedings{yu2019sparc,
  title={{SParC}: Cross-Domain Semantic Parsing in Context},
  author={Yu, Tao and Zhang, Rui and Yasunaga, Michihiro and Tan, Yi Chern and Lin, Xi Victoria and Li, Suyi and Er, Heyang and Li, Irene and Pang, Bo and Chen, Tao and others},
  booktitle={Proceedings of the 57th Annual Meeting of the Association for Computational Linguistics},
  pages={4511--4523},
  year={2019}
}

@inproceedings{yu2019cosql,
  title={{CoSQL}: A Conversational Text-to-{SQL} Challenge Towards Cross-Domain Natural Language Interfaces to Databases},
  author={Yu, Tao and Zhang, Rui and Er, Heyang and Li, Suyi and Xue, Eric and Pang, Bo and Lin, Xi Victoria and Tan, Yi Chern and Shi, Tianhao and Li, Zifan and others},
  booktitle={Proceedings of the 2019 Conference on Empirical Methods in Natural Language Processing},
  pages={1962--1979},
  year={2019}
}

@inproceedings{gan2021towards,
  title={Towards Robustness of Text-to-{SQL} Models Against Natural and Realistic Adversarial Table Perturbation},
  author={Gan, Xinyu and Chen, Jinchuan and Liu, Xuguang and Zheng, Binyuan and Wan, Yong and Song, Daxin and Lao, Bailin and Jiang, Minghua},
  booktitle={Proceedings of the 59th Annual Meeting of the Association for Computational Linguistics},
  pages={2007--2022},
  year={2021}
}

@inproceedings{chang2023dr,
  title={Dr.~{Spider}: A Diagnostic Evaluation Benchmark towards Text-to-{SQL} Robustness},
  author={Chang, Shuaichen and Xu, Jun and Sun, Tao and Gao, Peixiang and Lin, Zhenwen and Liang, Yan and Guo, Keji and others},
  booktitle={International Conference on Learning Representations},
  year={2023}
}

@inproceedings{lee2021kaggledbqa,
  title={{KaggleDBQA}: Realistic Evaluation of Text-to-{SQL} Parsers},
  author={Lee, Chia-Hsuan and Polozov, Oleksandr and Richardson, Matthew},
  booktitle={Proceedings of the 59th Annual Meeting of the Association for Computational Linguistics},
  pages={2261--2273},
  year={2021}
}

@inproceedings{liu2023divide,
  title={Divide and Prompt: Chain of Thought Prompting for Text-to-{SQL}},
  author={Liu, Xiaodan and Chen, Jiaxin and Shi, Jingfeng and Chen, Zhuo},
  booktitle={Findings of EMNLP},
  year={2023}
}

@article{lei2024spider2,
  title={Spider~2.0: Evaluating Language Models on Real-World Enterprise Text-to-{SQL} Workflows},
  author={Lei, Fangyu and Chen, Jixuan and Peng, Yiming and Li, Tongxin and Wang, Haoyang and Ge, Miao and others},
  journal={arXiv preprint arXiv:2411.07763},
  year={2024}
}

@misc{gartner2023db,
  title={Magic Quadrant for Cloud Database Management Systems},
  author={{Gartner Research}},
  year={2023},
  note={Accessed 2026}
}

@inproceedings{woods1973progress,
  title={Progress in Natural Language Understanding: An Application to Lunar Geology},
  author={Woods, William A},
  booktitle={Proceedings of the June 4--8, 1973, National Computer Conference and Exposition},
  pages={441--450},
  year={1973}
}

\end{document}